\documentclass{article}

\usepackage[final]{corl_2020} 

\usepackage[utf8]{inputenc} 
\usepackage[T1]{fontenc}    
\usepackage{hyperref}       
\usepackage{url}            
\usepackage{booktabs}       
\usepackage{amsfonts}       
\usepackage{nicefrac}       
\usepackage{microtype}      

\usepackage{multicol}
\usepackage{color}
\usepackage{amssymb}
\usepackage{amsmath}
\usepackage{amsthm}
\usepackage{tikz}
\usepackage{chngcntr}
\usepackage[makeroom]{cancel}
\usetikzlibrary{arrows,decorations.markings}

\usepackage{booktabs}
\usepackage{graphicx}
\usepackage{mathtools}
\usepackage{bbm}

\usepackage{caption}
\usepackage{subcaption}

\usepackage{algorithm2e}
\let\oldnl\nl
\newcommand{\nonl}{\renewcommand{\nl}{\let\nl\oldnl}}

\usepackage{indentfirst}

\newenvironment{tightlist}%
{\begin{list}{$\bullet$}{%
    \setlength{\topsep}{0in}
    \setlength{\partopsep}{0in}
    \setlength{\itemsep}{0in}
    \setlength{\parsep}{0in}
    \setlength{\leftmargin}{1.5em}
    \setlength{\rightmargin}{0in}
}
}%
{\end{list}
}

\newcommand{\secref}[1]{(\textsection \ref{#1})}

\newcommand{\eqnref}[1]{Equation~\ref{#1}}
\newcommand{\figref}[1]{Figure~\ref{#1}}

\newcommand{\tabref}[1]{Table~\ref{#1}}

\SetCommentSty{mycommfont}

\newtheorem{defn}{Definition}

\counterwithin*{thm}{section}

\def\thickhline{%
  \noalign{\ifnum0=`}\fi\hrule \@height \thickarrayrulewidth \futurelet
   \reserved@a\@xthickhline}
\def\@xthickhline{\ifx\reserved@a\thickhline
               \vskip\doublerulesep
               \vskip-\thickarrayrulewidth
             \fi
      \ifnum0=`{\fi}}

\newlength{\thickarrayrulewidth}
\DeclareMathOperator*{\argmax}{arg\!max}
\DeclareMathOperator*{\argmin}{arg\!min}

\usepackage{multirow}
\usepackage{geometry}
\usepackage{wrapfig}
\newcommand{\mdp}{{\sc mdp}}
\newcommand{\camp}{{\sc camp}}
\newcommand{\namo}{{\sc namo}}
\newcommand{\tamp}{{\sc tamp}}

\newcommand{\indep}{\perp\!\!\!\perp}
\newcommand{\Ex}{\mathop{\mathbb{E}}}

\newcommand{\C}{\mathcal{C}}

\renewcommand{\S}{\mathcal{S}}
\newcommand{\A}{\mathcal{A}}

\newcommand{\M}{\mathcal{M}}

\newcommand{\plan}{\textsc{Plan}}

\title{CAMPs: Learning Context-Specific Abstractions for Efficient Planning in Factored MDPs}

\author{
  Rohan Chitnis$^{\dagger *}$,
Tom Silver$^\dagger$\thanks{Equal contribution.}\ \ ,
Beomjoon Kim$^\mathsection$,
Leslie Pack Kaelbling$^\dagger$,
Tom\'as Lozano-P\'erez$^\dagger$\\
  $^\dagger$MIT Computer Science and Artificial Intelligence Laboratory \ $^\mathsection$KAIST Graduate School of AI\\
  \texttt{\{ronuchit, tslvr, lpk, tlp\}@mit.edu, beomjoon.kim@kaist.ac.kr}
}

\begin{document}

\maketitle

\begin{abstract}
Meta-planning, or learning to guide planning from experience, is a promising approach to improving the computational cost of planning. 
A general meta-planning strategy is to learn to impose constraints on the states considered and actions taken by the agent.
We observe that (1) imposing a constraint can induce \emph{context-specific independences} that render some aspects of the domain irrelevant, and (2) an agent can take advantage of this fact by imposing constraints \emph{on its own behavior}.
These observations lead us to propose the context-specific abstract Markov decision process (\camp{}), an abstraction of a factored \mdp{} that affords efficient planning. We then describe how to learn constraints to impose so the \camp{} optimizes a trade-off between rewards and computational cost.
Our experiments consider five planners across four domains, including robotic navigation among movable obstacles (\namo{}), robotic task and motion planning for sequential manipulation, and classical planning.
We find planning with learned \camp{}s to consistently outperform baselines, including Stilman's \namo{}-specific algorithm.
Video: \url{https://youtu.be/wTXt6djcAd4}
Code: \url{https://git.io/JTnf6}
\end{abstract}
\setcounter{footnote}{0}

\section{Introduction}
\label{sec:intro}

Online planning is a popular paradigm for sequential decision-making in robotics and beyond, but its practical application is limited by the computational burden of planning while performing a task.
In \emph{meta-planning},  the agent learns to guide planning efficiently and effectively based on its previous planning experience.
\emph{Learning to impose constraints} on the states considered and actions taken by an agent is a promising paradigm for meta-planning; it reduces the space of policies the agent must consider \cite{ltampkim,driess2020rss,wells2018ral}.
In contrast to (e.g., physical or kinematic) constraints beyond the agent's control, these constraints are imposed by the agent on itself for the sole purpose of efficient planning.

\begin{figure}[t]
\includegraphics[width=\textwidth]{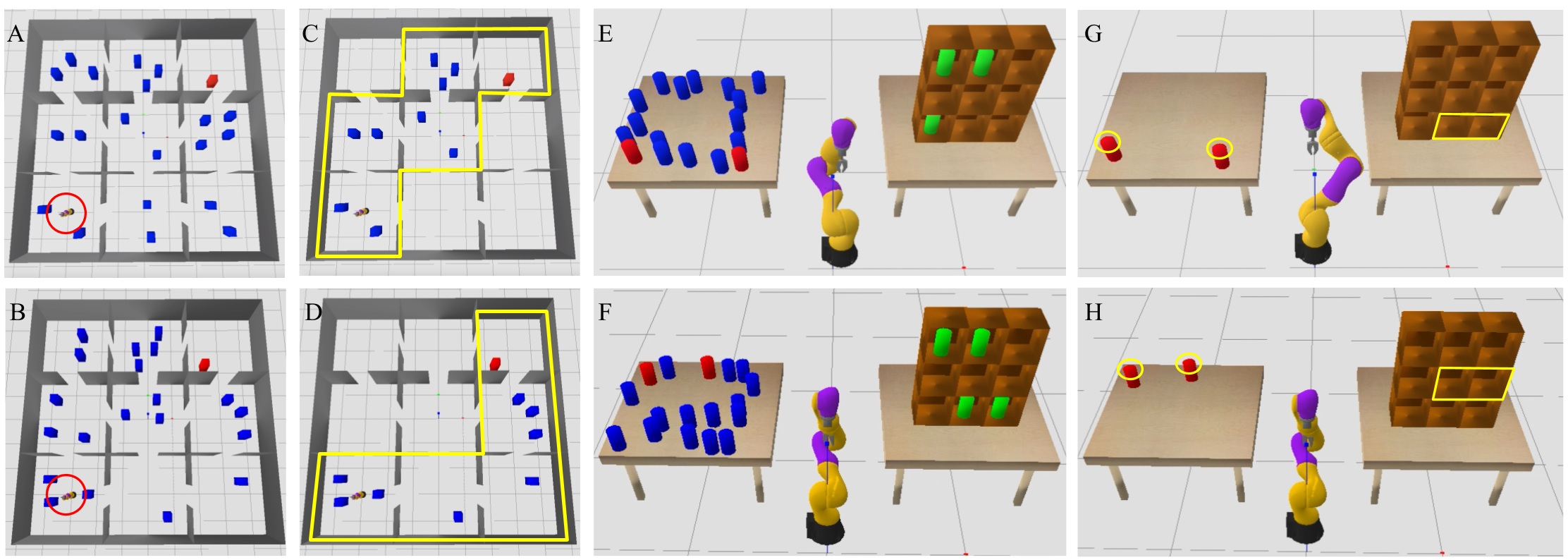}
\centering
\caption{
(A, B) In the \namo{} domain, the robot (red circle) must reach the red object, which requires navigating there while moving obstacles out of the way. Two sample problems are shown.
(C, D) If the robot constrains itself to stay within certain rooms (yellow), the obstacles in other rooms become irrelevant.
(E, F) In the sequential manipulation domain, the robot must put the red objects into bins.
(G, H) If the robot constrains itself to \emph{top} grasps, the blue objects become irrelevant. Similarly, if the robot constrains target placements to certain bins (yellow), the green objects become irrelevant.
}
\label{fig:teaser}
\end{figure}

Beyond reducing the space of policies, imposing constraints can improve planning efficiency in another important way. In factored domains, where the states and actions decompose into variables,
imposing constraints can induce \emph{context-specific independences} (CSIs) \cite{boutilier1996context} that render some variables irrelevant.
For example, consider the two navigation among movable obstacles (\namo{}) problems in Figure \ref{fig:teaser}A--B. Imposing a constraint that forbids certain rooms induces CSIs between the robot's position and that of all obstacles in those forbidden rooms.
Consequently, these obstacles can be ignored, as in Figure \ref{fig:teaser}C--D.
A planning problem with a context imposed\footnote{We henceforth use \emph{context} as a synonym for \emph{constraint}.} and the resulting irrelevant variables removed constitutes an \emph{abstraction} of the original problem \cite{liabstractions,konidarisabstractions}.
Adopting the Markov decision process (\mdp{}) formalism, we refer to this as a \emph{context-specific abstract \mdp{}} (\camp{}).

Planning in a \camp{} is often more efficient than planning in the original \mdp{}, but may abstract away important details of the environment, leading to a suboptimal policy. 
Practically speaking, we are often interested in a trade-off: we would like our planners to produce highly rewarding behavior, but not be too computationally burdensome; we are willing to sacrifice optimality, and in the case of goal-based tasks, even soundness and completeness, to maximize this trade-off in expectation.
In this work, we propose a learning-based approach to maximize this trade-off.
Given a set of training tasks with a shared transition model and factored states and actions, we first approximate the set of CSIs present in these tasks. 
We then train a \emph{context selector}, which predicts a context that should be imposed for a given task.
At test time, given a novel task, we use the learned context selector and CSIs to induce a \camp{}, which we then use to plan.
This overall pipeline is summarized in Figure \ref{fig:learning}.

Our approach rests on the premise that predicting contexts to impose is easier, and generalizes better, than learning a reactive policy. Intuitively, the burden on reactive policy learning is higher, as the policy must exactly carve out a specific, good path through transition space, whereas an imposed context must only carve out a region of transition space that includes at least one good path.

In experiments, we consider four domains, including robotic \namo{} and sequential manipulation, that collectively exhibit discrete and continuous states and actions, relational states, sparse rewards, stochastic transitions, and long planning horizons.
To evaluate the generality of \camp{}s, we consider multiple planners, including Monte Carlo tree search \cite{mcts}, FastDownward \cite{fd}, and a task and motion planner \cite{srivastava2014combined}.
Our results suggest that planning with learned \camp{}s strikes a strong balance between pure planning and pure policy learning \cite{moerl2020think}.
In the \namo{} domain, we also find that \camp{}s with a generic task and motion planner outperform Stilman's \namo{}-specific algorithm \cite{namo}.
We conclude that \camp{}s offer a promising path toward fast, effective planning in large and challenging domains.
\section{Preliminaries}

We introduce notation and background formalism \secref{subsec:csifmdp}, and then present a problem definition \secref{subsec:problemformulation}.


\subsection{Context-Specific Independence in Factored Markov Decision Processes}
\label{subsec:csifmdp}
A Markov decision process (\mdp{}) is given by $(\S, \A, T, R, H)$, with: state space $\S$; action space $\A$; transition model $T(s_t, a_t, s_{t+1}) = P(S_{t+1}=s_{t+1} \mid S_{t}=s_t, A_t=a_t)$ where $s_t, s_{t+1} \in \S$, $a_t \in \A$, and $S_t$, $A_t$ are random variables denoting the state and action taken at time $t$; reward function $R(s_t) = r_t \in \mathbb{R}$; and horizon $H$.\footnote{We say the reward is a function of only $s_t$ for simplicity of notation; this is not critical to our method.}
The solution to an \mdp{} is a policy $\pi: \S \to \A$, a mapping from states to actions, that maximizes the expected sum of rewards with respect to the transitions.

We focus on \emph{factored} \mdp{}s~\cite{guestrin2003efficient}, where each state variable $S$ is factored into $n$ variables $\{S^1, \ldots, S^n\}$, where $S^i$ has domain $\mathcal{S}^i$. A state $s \in \mathcal{S}$ is then an assignment $s = [s^1, \ldots, s^n]$ with $s^i \in \mathcal{S}^i$; thus, $\mathcal{S} \subseteq \mathcal{S}^1 \times \ldots \times \mathcal{S}^n$.
Actions are similarly factored into $m$ variables $\{A^1, \ldots, A^m\}$ with domains $\mathcal{A}^i$ so that $a \in \mathcal{A}$ is an assignment $[a^1, \ldots, a^m]$ with $a^i \in \mathcal{A}^i$; thus, $\mathcal{A} \subseteq \mathcal{A}^1 \times \ldots \times \mathcal{A}^m$.
The reward function for a factored \mdp{} is defined in terms of a subset of state variables $S_{\text{rew}} \subseteq \{S^1, \ldots, S^n\}$, which we call the \emph{reward variables}.
Let $V = \{S^1, \ldots, S^n\} \cup \{ A^1, \ldots, A^m \}$ denote all state and action variables together.
Variable domains may be discrete or continuous for both states and actions.


Following \cite{boutilier1996context}, we define a \emph{context} as a pair $(C,\C)$, where $C \subseteq V$ is some subset of state and action variables, and  $\C$ is a space of possible joint assignments.
A state-action pair $(s, a)$ is \emph{in the context} $(C,\C)$ when its joint assignment of variables in $C$ is present in $\C$. Two variables $X, Y \subseteq V \setminus C$ are \emph{contextually independent} under $(C, \C)$ if $P\left(X \mid Y, C=c\right) = P(X \mid C=c)\ \forall\ c \in \C$,
in which case we write $X \indep Y \mid (C,\C)$.
This relation is called a \emph{context-specific independence} (CSI).
In this paper, we explore how CSIs can be automatically identified and exploited for planning in factored \mdp{}s.


\begin{figure}[t]
\includegraphics[width=\textwidth]{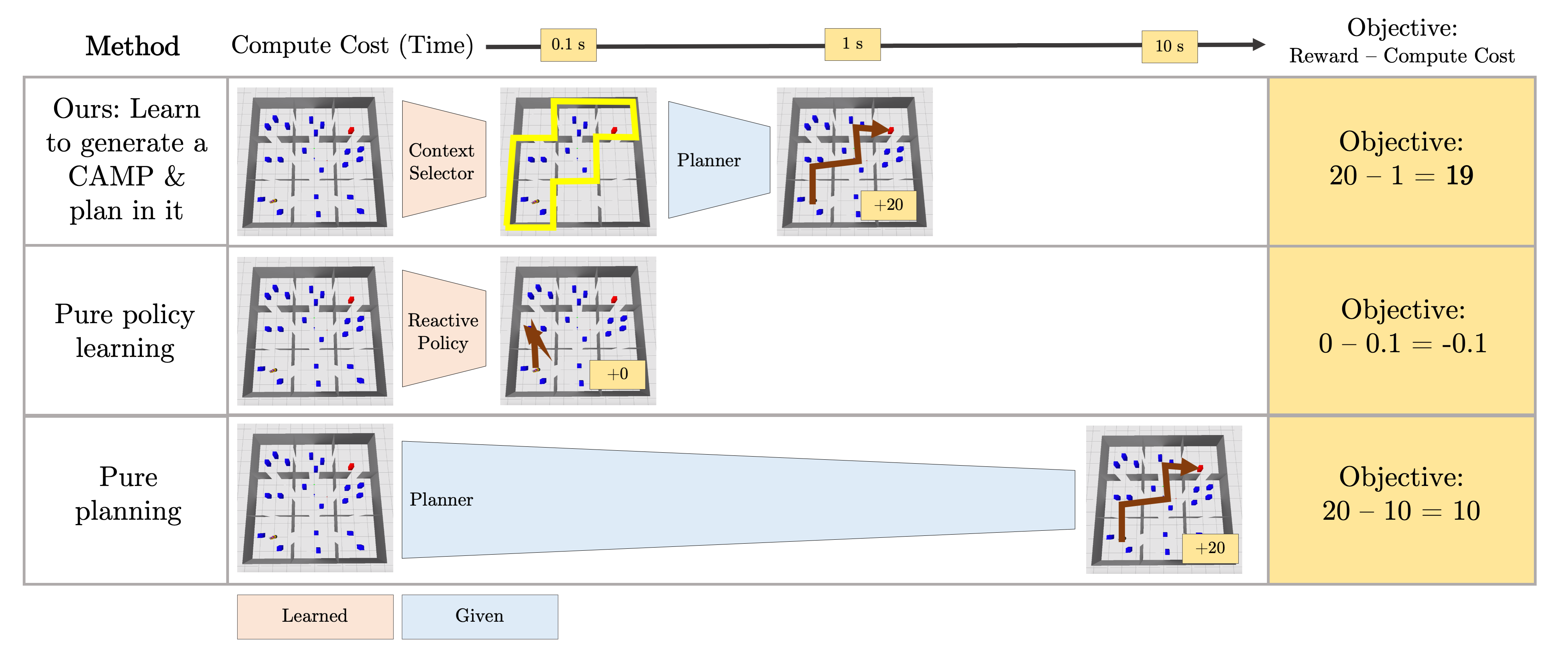}
\centering
\caption{
Three approaches to solving an \mdp{}.
Given a task, our approach (top row) applies its learned context selector to generate a \camp{}, then plans in this \camp{} to get a policy.
Our approach often achieves higher reward than pure policy learning (middle row), and lower computational cost than pure planning (bottom row), leading to a good objective value (right; uses $\lambda=1$ in \eqnref{eq:objective}).
}
\label{fig:learning}
\end{figure}

\subsection{Problem Formulation}
\label{subsec:problemformulation}
A \emph{task} is a pair of initial
state and reward function, denoted $\omega = (s_0,R)$.
We are given a set of $N$ training tasks, $W_{\text{train}} = \{\omega^{(i)}\}_{i=1}^{N}$, and a test task, $\omega_{\text{test}}$, all of which are drawn from some unseen distribution $P(\mathcal{W})$. All tasks share the same factored state space $\S$, factored action space $\A$, transition model $T$, and horizon $H$; therefore, each task induces a factored \mdp{}, denoted $\mathcal{M}_{\omega}$. Each task is parameterized by a feature vector, denoted $\theta_\omega = \phi(s_0,R)$, with featurizer $\phi$. For instance, in our robotic \namo{} domain, $\phi$ gives a top-down image of the initial scene (which also implicitly describes the goal). The agent interacts with $T$ and $R$ as black boxes; it does not know their analytical representations or causal structure. 
We also assume a black-box \mdp{} solver, \plan{}, which takes as input an \mdp{} $\mathcal{M}$ and a current state $s$, and returns a next action: $a = \plan{}(\mathcal{M}, s)$.\footnote{Planners generally return a policy or a sequence of actions; we suppose that the planner is called at every timestep to simplify exposition. In our experiments, we replan in domains that have stochastic transitions.}

Before being presented with the test task, the agent may first interact with the training tasks $W_{\text{train}}$, perhaps compiling useful knowledge that it can deploy at test time, such as a task-conditioned policy.
Then, it is given the test task $\omega_{\text{test}} = (s_{0,\text{test}}, R_{\text{test}})$, and its goal is to \emph{efficiently} produce actions that accrue high cumulative reward in the test \mdp{} $\mathcal{M}_{\omega_{\text{test}}}$.
We formalize this trade-off via the objective:
\begin{equation}
    \label{eq:objective}
     J(\pi, \omega)= \Ex\left[\sum_{t=0}^{H} R(s_t) - \lambda \cdot \textsc{ComputeCost}(\pi, s_t)\right],
\end{equation}
where $\omega = (s_0, R)$, $\textsc{ComputeCost}(\pi, s) \ge 0$ denotes the cost (e.g., wall-clock time) of evaluating the policy $\pi(s)$, $\lambda \ge 0$ is a trade-off parameter, and the expectation is over stochasticity in the transitions. Note that \textsc{ComputeCost} includes the cost of both computation performed before the agent starts acting and any computation that might be performed on each timestep after the first.

We seek to find $\pi_{\text{test}} = \argmax_\pi J(\pi, \omega_{\text{test}})$.
One possible approach is to call $\plan{}$ on the full test \mdp{}, that is, $\pi_{\text{test}}(s) = \plan(\mathcal{M}_{\omega_{\text{test}}}, s)$.
This method would yield high rewards, but it may also incur a large \textsc{ComputeCost}.
Another possibility is to learn (at training time) and transfer (to test time) a task-conditioned reactive policy; this can have low \textsc{ComputeCost} at test time, but perhaps at the expense of rewards if the policy fails to generalize well to the test task (\figref{fig:learning}).
\section{Context-Specific Abstract Markov Decision Processes (CAMPs)}
\label{sec:representation}

The objective formulated in Equation \ref{eq:objective} trades off the computational cost of planning with the resulting rewards.
In this section, we present an approach to optimizing this trade-off that lies between the two extremes of pure planning and pure policy learning \cite{moerl2020think}.
Rather than planning in the full test task, we propose to \emph{learn} to generate an abstraction \cite{liabstractions,konidarisabstractions} of the test task, in which we can plan efficiently.

An \emph{abstraction} over state space $\S$ and action space $\A$ is a pair of functions $(\sigma, \tau)$, with $\sigma : \S \mapsto \S'$ and $\tau : \A \mapsto \A'$, where $\S'$ and $\A'$ are \emph{abstract} state and action spaces.
We are specifically interested in abstractions that are projections:
$\sigma([s^1, s^2, \ldots, s^n]) = [s^{i_1}, s^{i_2}, \ldots, s^{i_{n'}}] \text{ and } \tau([a^1, a^2, \ldots, a^m]) = [a^{j_1}, a^{j_2}, \ldots, a^{j_{m'}}]$. This has the effect of dropping $n-n'$ state variables and $m-m'$ action variables; the $i$ and $j$ superscripts refer respectively to the state and action variables that are not dropped.



The \emph{relevant-variables projection} is a simple projective abstraction that has been studied in prior work (under different names)~\cite{baum2012proximity,gardiolthesis,boutilier1997correlated}. It drops all irrelevant variables, in the following sense:

\begin{defn}[Variable relevance]
\label{defn:relevance}
Given a factored \mdp{} with variables $V$ and reward variables $S_{\text{rew}}$, any $V^i \in V$ is \emph{relevant} iff $\exists\ V^j \in S_{\text{rew}}$, $t \in \{0, \ldots, H\}$, and $t' \in \{t+1, \ldots, H\}$ s.t. $V_t^i \not\indep V_{t'}^j$.
\end{defn}

Intuitively, a variable is relevant if there is \emph{any} possibility that its value at some timestep will have an eventual influence, directly or indirectly, on the value of the reward.
Unfortunately, as identified by \citet{baum2012proximity}, relevance is often too strong of a property for the relevant-variables projection to yield meaningful improvements in practice --- most variables typically have \emph{some} way of influencing the reward, under \emph{some} sequence of actions taken by the agent.
In search of greater flexibility, we now define a generalization of variable relevance that is conditioned on a particular context \secref{subsec:csifmdp}.
\begin{defn}[Context-specific variable relevance]
\label{defn:cs-relevance}
Given a context $(C,\C)$ and a factored \mdp{} with variables $V$ and reward variables $S_{\text{rew}}$, any $V^i \in V \setminus C$ is \emph{relevant in the context} $(C,\C)$ iff $\exists\ V^j \in S_{\text{rew}}$, $t \in \{0, \ldots, H\}$, and $t' \in \{t+1, \ldots, H\}$ s.t. $V_t^i \not\indep V_{t'}^j \mid (C,\C)$.
\end{defn}
Each possible context $(C,\C)$ induces a projection that drops variables which are irrelevant in $(C,\C)$; let $\text{proj}_{C,\C}$ denote this abstraction. We now define a \camp{}, an abstract \mdp{} associated with $\text{proj}_{C,\C}$.
\begin{defn}[Context-Specific Abstract \mdp{} (\camp{})]
\label{defn:camp}
Consider an \mdp{} $\M = (\S, \A, T, R, H)$ and a context $(C,\C)$. Let $\text{proj}_{C,\C} = (\sigma, \tau)$ with right inverses $(\sigma^{-1}, \tau^{-1})$.
Let $\bot$ be a new sink state, such that $\bot \not\in \sigma(\S)$. The context-specific abstract \mdp{}, $\M'$, for $\M$ and $(C,\C)$ is $(\sigma(\S) \cup \{ \bot \}, \tau(\A), T', R', H)$, where $T'$ and $R'$ are defined as follows: $\forall s'_t, s'_{t+1} \in \sigma(\S), a'_t \in \tau(\A),$
\begin{enumerate}
    \item $T'(\bot, a_t', \bot) = 1$
    \item $T'(s_t', a_t', \bot) = 1$ if $(\sigma^{-1}(s_t'), \tau^{-1}(a_t'))$ is not in the context;
    \item $T'(s_t', a_t', s'_{t+1}) = T(\sigma^{-1}(s_t'), \tau^{-1}(a_t'), \sigma^{-1}(s'_{t+1}))$ if $(\sigma^{-1}(s_t'), \tau^{-1}(a_t'))$ is in the context;
    \item $R'(\bot) = -\infty$
    \item $R'(s_t') = R(\sigma^{-1}(s_t'))$
\end{enumerate}
We may also say that $\M'$ is $\M$ \emph{with context $(C, \C)$ imposed}.
\end{defn}

Intuitively, a \camp{} imposes a projective abstraction that drops the variables that are irrelevant \emph{under the given context}, and also imposes that any transition in violation of the context leads the agent to $\bot$, an absorbing sink state with reward $-\infty$. In practice, the right inverses $\sigma^{-1}$ and $\tau^{-1}$ can be obtained by assigning arbitrary values to the dropped variables; the choice of value is inconsequential by Definition \ref{defn:cs-relevance}. For a graphical example of a \camp{}, see Appendix \ref{sec:campgraphicalexample}.


A \camp{} is usually \emph{not} optimality-preserving, because the context restricts the agent to a subregion of the state and action space~\cite{abel2019value}. However, context-specific relevance is much weaker than relevance: it only requires a variable to be relevant under the given context. For example, to a robot operating in a home, the weather outside is irrelevant as long as it remains in the context of staying indoors.

\camp{}s offer a way to solve the test task \secref{subsec:problemformulation} that lies between the extremes of pure planning and pure policy learning. Namely, given a test \mdp{} $\M_{\omega_{\text{test}}}$, we select a context, compute the relevant variables under that context via backward induction, generate the \camp{}, and finally plan in this \camp{} to obtain a policy $\pi_{\text{test}}$ for $\M_{\omega_{\text{test}}}$.
We have therefore reduced the problem of optimizing $J(\pi, \omega_{\text{test}})$ to that of determining the best context $(C, \C)$ to impose.
To address this issue, we now turn to learning.

\section{Learning to Generate CAMPs}

\begin{figure}[t]
\includegraphics[width=\textwidth]{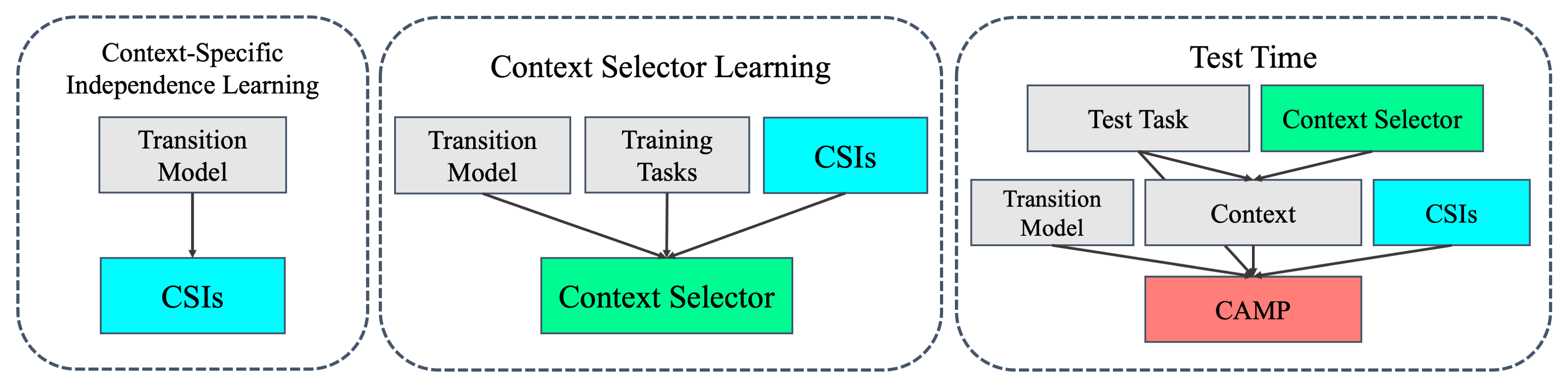}
\centering
\caption{Data-flow diagram for our method during learning and test time.
(Left) Approximate context-specific independence (CSI) learning derives the relevant state and action variables under each context.
(Middle) A context selector is learned by optimizing the objective on the training tasks.
The transition model and approximate CSIs are used to evaluate the objective.
(Right) Given a test task, the agent selects a context to impose using the learned context selector.
The relevant variables for this context are calculated from the learned CSIs.
From the context, relevant variables, test task, and transition model, the agent derives a \camp{}, and can plan in it to obtain a policy for the test task.
}
\label{fig:blockdiagram}
\end{figure}

We now have the ability to generate a context-specific abstract \mdp{} (\camp{}) when given a context and the associated context-specific independences.
However, contexts and their associated independences are \emph{not} provided in our problem.
In this section, we describe how to learn approximate context-specific independences and a context selector, for use at test time.
\figref{fig:blockdiagram} gives a data-flow diagram.

\subsection{Approximating the Context-Specific Independences}
\label{subsec:csilearning}

Recall that the agent is given \mdp{}s with factored states and variables, but only query access to the (shared) transition model.
For example, the transition model may be a black-box physics simulator, as in two of our experimental domains.
In order to approximately determine the context-specific independences that are latent within a factored \mdp{}, we propose a sample-based procedure.
Given a context, the algorithm examines each pair of state or action variables $(V^i, V^j)$ and tests for empirical dependence, that is, whether any sampled value of $V_t^i$ induces a change in the distribution of $V_{t+1}^j$, conditioned on the sampled values of the remaining variables. For full pseudocode, see Appendix \ref{sec:csipseudocode}.

The runtime of this algorithm depends on the size of the domain and the number of samples used to test dependence.
In theory, the number of samples required to identify all independences could be arbitrarily large.
In practice, for the tasks we considered in our experiments, including robotic manipulation and \namo{}, we found this algorithm to be sufficient for detecting a useful set of context-specific independences. Moreover, our method is robust to errors in the discovered independences, which in the worst case will simply exclude some candidate abstractions from consideration.



\textbf{Where Does the Space of Contexts Come From?} 
Our approximate algorithm allows us to estimate independences \emph{given a context};
this raises the question of which contexts should be evaluated.
We propose a simple method for deriving a space of possible contexts that works well across our varied experimental domains.
From the set of variables $V$, we consider conjunctions and disjunctions up to some length (a hyperparameter), excluding any terms whose involved variables have joint domain size less than some threshold (another hyperparameter).
Note that for any finite threshold, this procedure immediately excludes contexts involving continuous variables.
While this family of contexts has the benefit of being fairly general to describe, we emphasize that other choices, e.g., more domain-specific context families, may be used as well \cite{ltampkim,driess2020deep,carpentier2017learning}.
Importantly, the space of considered contexts should always include the trivial universal context so that \camp{}s can reduce to planning in the original problem when no useful abstractions are available.

\subsection{Learning the Context Selector}
\label{subsec:selectorlearning}

The performance of a \camp{} depends entirely on the selected context; if the context constrains the agent to a poor region of plan space, or induces independences that make important variables irrelevant, the resulting policy could get very low rewards.
However, if the context is selected judiciously, the \camp{} may exhibit substantial efficiency gains with minor impact on rewards.

We now describe an algorithm for learning to select a context that optimizes the objective (\eqnref{eq:objective}).
Pseudocode is presented in Appendix \ref{sec:selectorpseudocode}.
Given each training task $\omega^{(i)} \in W_{\text{train}}$, we first identify the best possible context $(C^{(i)}, \C^{(i)})^*$ according to the objective in Equation \ref{eq:objective}.
This process sets up a supervised multiclass classification problem that maps the featurized representation of a task $\theta_{\omega^{(i)}}$ to the best context $(C^{(i)}, \C^{(i)})^*$ to impose on that task.
We solve this classification problem by training a neural network with cross-entropy loss, resulting in a context selector $f_{\alpha}(\theta_\omega) = (C, \C)^*$, where $\alpha$ denotes the parameters of the neural network. At test time, we choose a context by calling $f_{\alpha}(\theta_{\omega_{\text{test}}})$, generate the associated \camp{}, and plan in this \camp{} to efficiently obtain a policy for the test task.

\section{Experiments and Results}

Our experiments aim to answer the following key questions:
\begin{tightlist}
    \item How does planning with learned \camp{}s compare to pure planning and pure policy learning across a varied set of domains, both discrete and continuous? \secref{subsec:domains}, \secref{subsec:results}
    \item To what extent is the performance of \camp{}s planner-agnostic? \secref{subsec:domains}, \secref{subsec:results}, (Appendix \ref{subsec:offline})
    \item How does the performance of \camp{}s vary with the choice of $\lambda$ (Equation \ref{eq:objective})? \secref{subsec:lambda}
    \item How does the performance of \camp{}s vary with the number of training tasks? (Appendix \ref{subsec:numenvs})
\end{tightlist}

We overview the experimental setup in \secref{subsec:domains} and \secref{subsec:methods}, and provide details in Appendix \ref{subsec:details}. Then, we present our main results in \secref{subsec:results}, with additional results in \secref{subsec:lambda} and Appendix \ref{subsec:numenvs}--\ref{subsec:offline}.

\subsection{Domains and Planners}
\label{subsec:domains}

We consider four domains and five planners (four online, one offline). Full details are in Appendix \ref{sec:domplandetails}.

\textbf{Domain D1: Gridworld.} A maze-style gridworld in which the agent must navigate across rooms to reach a goal location, while avoiding or destroying stochastically moving obstacles.
Task features are top-down images of the maze layout. The state is a vector of the current position and room of each obstacle, the agent, and the goal. The actions are moving up, down, left, right; and destroying each obstacle.
For planning in this domain, we consider Monte Carlo tree search (\textbf{MCTS}), breadth-first graph search with replanning (\textbf{BFSReplan}), and value iteration (\textbf{VI}). VI results are in Appendix \ref{subsec:offline}.

\textbf{Domain D2: Classical planning.} A deterministic dinner-making domain written in PDDL \cite{pddl}, with three different possible meals to make.
Preparing each meal requires a different number of actions.
The relative rewards for making each meal are the only source of variation between tasks; task features are simply these rewards. States are binary vectors describing which logical fluents hold true, and actions are logical operators.
We use an off-the-shelf classical planner (Fast-Downward~\cite{fd}).

\textbf{Domain D3: Robotic navigation among movable obstacles (\namo{})}, simulated in PyBullet~\cite{pybullet}.
Task features are overhead images. 
The state is a vector of the current pose of each object and the robot, and the robot's current room. The actions are moving the robot to a target pose, and clearing an object in front of the robot.
We use a state-of-the-art \tamp{} planner~\cite{srivastava2014combined}, which is \emph{not} \namo{}-specific.

\textbf{Domain D4: Robotic sequential manipulation}, simulated in PyBullet~\cite{pybullet}.
Task features are a vector of the object radii and occupied bins. The state is a vector of the current pose of each object, the grasp style used by the robot, and the current held object (if any). The actions are moving the robot to a target base pose and grasping at a target gripper pose, and moving the robot to a target base pose and placing at a target placement pose.
The planner for this task is the same as in \namo{}.

\subsection{Methods and Baselines}
\label{subsec:methods}
We consider the following methods:

\begin{tightlist}
    \item \camp{}. Our full method.
    \item \camp{} ablation. An ablation of our full method in which the \camp{} only sends the agent to a sink state for context violation, but does \emph{not} project away irrelevant variables.
    \item Pure planning. This baseline does not use the training tasks, and just solves the full test task.
    \item Plan transfer. This baseline solves each training task to obtain a plan, and at test time picks actions via majority vote across the training task plans.
    \item Policy learning. This baseline solves each training task to obtain a plan, then trains a state-conditioned neural network policy to imitate the resulting dataset of state-action trajectories, using supervised learning. This policy is used directly to choose actions at test time.
    \item Task-conditioned policy learning. This baseline is the same as policy learning, but the neural network also receives as input the features of the task, in addition to the current state.
    \item (Domain D3 only) Stilman's planning algorithm~\cite{namo} for \namo{} problems, named ResolveSpatialConstraints, which attempts to find a feasible path to a target location by first finding feasible paths to any obstructing objects and moving them out of the way.
\end{tightlist}

In all our domains, every variable is relevant under \emph{no} context. For this reason, the pure planning baseline can also be understood as an ablation of \camp{} that does not account for contexts.

\subsection{Main Results}
\label{subsec:results}

\begin{figure}[t]
\includegraphics[width=\textwidth]{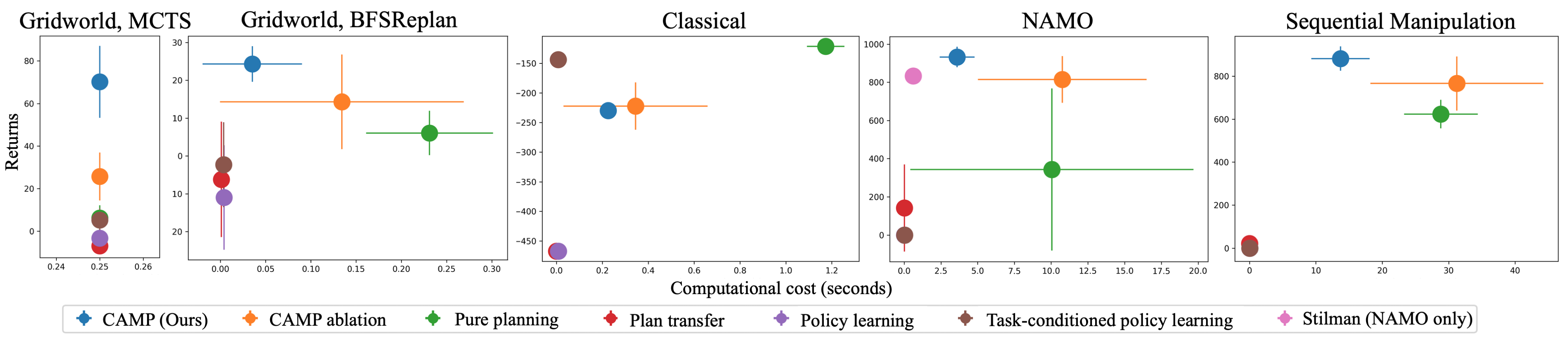}
\centering
\caption{
Mean returns versus computation time on the test tasks, for all domains and methods. All points report an average over 10 independent runs of training and evaluation, with lines showing per-axis standard deviations. \camp{}s generally provide a better trade-off than the baselines: the blue points are usually higher than pure policy learning (\camp{}s accrue more reward) and to the left of pure planning (\camp{}s are more efficient). For the left-most plot, only the returns vary because MCTS is an anytime algorithm, so we run it with a fixed timeout. See \secref{subsec:results} for discussion on these results.}
\label{fig:plots2d}
\end{figure}

\figref{fig:plots2d} plots the mean returns versus computation time on the test tasks, for all domains and methods. \tabref{tab:mainresults} in Appendix \ref{sec:objectivetable} shows the corresponding objective values (\eqnref{eq:objective}). All results report an average over 10 independent runs of context selector training and test task evaluation.

\emph{Discussion.}
\camp{}s outperform every baseline in all but Domain D2 (classical planning).
\camp{}s fare better than task-conditioned policy learning because the latter fails to generalize from training tasks to test tasks.
This failure manifests in low test task rewards, and in a substantial difference between the training and test objective values.
In classical planning, however, policy learning outperforms \camp{}s; both achieve high task rewards, but the policy is faster to execute.
This is because this domain involves little variation between instances, in stark contrast to the other domains.
In Appendix \ref{subsec:numenvs}, we unpack this result further by analyzing performance versus number of training tasks.

Another clear conclusion from the main results is that \camp{}s outperform pure planning across all experiments, consistently achieving lower computational costs.
In several cases, including \namo{} and manipulation, \camp{}s also achieve higher \emph{rewards} than pure planning does, since the latter sometimes hits the 60-second timeout before discovering the superior plans found very quickly by \camp{}s.

Results for the \camp{} ablation show that imposing contexts alone provides clear benefits, focusing the planner on a promising region of the search space.
This result is consistent with prior work showing that learning to impose constraints reduces planning costs \cite{ltampkim,driess2020rss,wells2018ral}. However, the difference between \camp{} and the ablation shows that dropping irrelevant variables provides even greater benefits.

A final observation is that \camp{}s perform comparably to Stilman's \namo{} algorithm~\cite{namo}.
This is notable because Stilman's algorithm employs \namo{}-specific assumptions, whereas the planner we use does not; in fact, we can see that the pure planner is strongly outperformed by Stilman's algorithm.
In our method, the context selector learns to constrain the robot to stay in emptier rooms, meaning it must move comparatively few objects out of the way. This leads to efficient planning, making the computational cost of \camp{}s almost as good as that of Stilman's algorithm; additionally, it leads \camp{}s to obtain higher rewards than Stilman's algorithm, because it often reaches the goal faster.
The \namo{} results also show that \camp{}s are able to learn to impose useful contexts even when there are multiple good options, e.g., multiple ``room paths'' with similar numbers of obstacles (\figref{fig:teaser}).

\subsection{Performance as a Function of $\lambda$}
\label{subsec:lambda}
The plots on the right illustrate how the objective value (left) and returns (right) accrued by the \camp{} policy vary as a function of $\lambda$, in Domain D2 (classical planning).
\begin{wrapfigure}{r}{0.5\textwidth}
\includegraphics[width=0.49\textwidth]{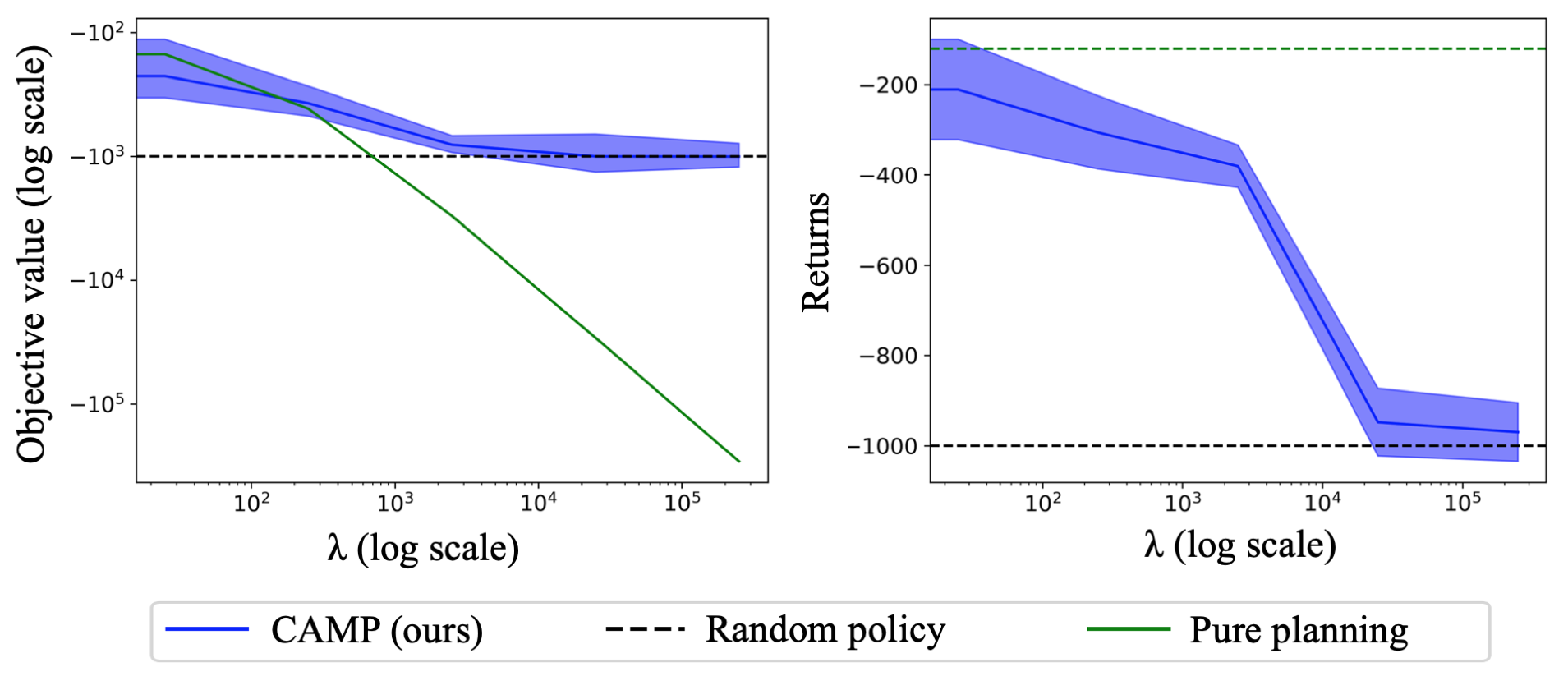}
\centering
\end{wrapfigure}
The right figure shows the returns from \camp{} interpolate between those obtained by pure planning (when $\lambda=0$, the agent is okay with spending a long time planning out its actions) and those obtained by a random policy (when $\lambda \to \infty$, the agent spends as little time as possible choosing actions). The green line is dashed because pure planning does not use $\lambda$, so its returns are unaffected by the value of $\lambda$. The left figure (note the log-scale $y$-axis) shows objective values. We see that \camp{} never suffers a lower objective value than that of a random policy, while pure planning drops down greatly as $\lambda$ increases. This is because as $\lambda \to \infty$, our context selector learns to choose contexts that induce very little planning (but get low returns).
\section{Related Work}
Our work falls under the broad research theme of learning to make planners more efficient using past planning experience. A fundamental question is deciding what to predict; for instance, it is common to learn a policy and/or value function from planning experience~\cite{alphago,ltampkim2,kim19learning, ltampchitnis}.
In contrast, we learn to predict \emph{contexts}.
Recent work leverages a given set of contexts to represent planning problem instances in ``score space''~\cite{ltampkim}, but does not consider the resulting CSIs, which we showed experimentally to yield large performance improvements.
Other methods predict the feasibility of task plans or motion plans~\cite{driess2020rss,wells2018ral,driess2020deep}, which can also be seen as learning constraints on the search space. These methods can be readily incorporated into the \camp{} framework.

We have formalized \camp{}s as a particular class of \mdp{} abstractions.
There is a long line of work on deriving abstractions for \mdp{}s, much of it motivated by the prospect of faster planning or more sample-efficient reinforcement learning 
\cite{liabstractions,gardiolthesis,jong2005state,steinkraus2005solving,abel2017near}.
One common technique is to \emph{aggregate} states and actions into equivalence classes \cite{bean1987aggregation,bertsekas1988adaptive,konidarisabstractions}, a generalization of our notion of projective abstractions.
Other work has learned to select abstractions~\cite{jiang2015abstraction,konidaris2016constructing};
a key benefit of \camp{}s is that the contexts induce a structured hypothesis space of abstractions that greatly improve planning efficiency.

\camp{}s identify and exploit CSIs \cite{boutilier1996context} in factored planning problems.
In graphical models, CSIs can be similarly used to speed up inference \cite{zhang1999role,poole2013context,domshlak2004efficient}.
Stochastic Planning using Decision Diagrams (SPUDD) is a method that adapts these insights for planning with CSIs \cite{spudd}.
These insights are orthogonal to \camp{}s, but could be integrated to yield further efficiencies.
SPUDD is a pure planning approach that considers \emph{all} contexts, whereas we \emph{learn} a context selector that induces abstractions.
\section{Conclusion}

In this work, we have presented a method for learning to generate context-specific abstractions of \mdp{}s, achieving more efficient planning while retaining high rewards.
There are several clear directions for future work.
On the learning side, one interesting question is whether factorizations of initially unfactored \mdp{}s can be automatically discovered in a way that leads to useful \camp{}s.
Another direction to pursue is learning the task featurizer $\phi$, which we assumed to be given in our problem formulation.
Following \cite{ltampkim}, it could also be useful to extend the methods we have presented here so that multiple contexts can be imposed in succession at test time, using the performance of previous contexts to inform the choice of future ones.
However, note that such a method would lead to an increase in computational cost, possibly to the detriment of the overall objective we formulated.




\clearpage
\acknowledgments{We would like to thank Kelsey Allen for valuable comments on an initial draft.
We gratefully acknowledge support from NSF grant 1723381; from AFOSR grant FA9550-17-1-0165; from ONR grant N00014-18-1-2847; from the Honda Research Institute; from MIT-IBM Watson Lab; and from SUTD Temasek Laboratories.
Rohan and Tom are supported by
NSF Graduate Research Fellowships. Any opinions, findings,
and conclusions or recommendations expressed in this material are those of the authors and do not necessarily reflect the
views of our sponsors.}


\bibliography{references}  

\begin{thebibliography}{39}
\providecommand{\natexlab}[1]{#1}
\providecommand{\url}[1]{\texttt{#1}}
\expandafter\ifx\csname urlstyle\endcsname\relax
  \providecommand{\doi}[1]{doi: #1}\else
  \providecommand{\doi}{doi: \begingroup \urlstyle{rm}\Url}\fi

\bibitem[Kim et~al.(2017)Kim, Kaelbling, and Lozano-P{\'e}rez]{ltampkim}
B.~Kim, L.~P. Kaelbling, and T.~Lozano-P{\'e}rez.
\newblock Learning to guide task and motion planning using score-space
  representation.
\newblock In \emph{Robotics and Automation (ICRA), 2017 IEEE International
  Conference on}, pages 2810--2817. IEEE, 2017.

\bibitem[Driess et~al.(2020)Driess, Ha, and Toussaint]{driess2020rss}
D.~Driess, J.-S. Ha, and M.~Toussaint.
\newblock Deep visual reasoning: Learning to predict action sequences for task
  and motion planning from an initial scene image.
\newblock In \emph{Proc{.} of Robotics: Science and Systems (R:SS)}, 2020.

\bibitem[Wells et~al.(2018)Wells, Dantam, Shrivastava, and
  Kavraki]{wells2018ral}
A.~M. Wells, N.~T. Dantam, A.~Shrivastava, and L.~E. Kavraki.
\newblock Learning feasibility for task and motion planning in tabletop
  environments.
\newblock \emph{IEEE Robots and Automation Letters}, 2018.

\bibitem[Boutilier et~al.(1996)Boutilier, Friedman, Goldszmidt, and
  Koller]{boutilier1996context}
C.~Boutilier, N.~Friedman, M.~Goldszmidt, and D.~Koller.
\newblock Context-specific independence in bayesian networks.
\newblock In \emph{Proceedings of the Twelfth conference on Uncertainty in
  artificial intelligence}. Morgan Kaufmann Publishers Inc., 1996.

\bibitem[Li et~al.(2006)Li, Walsh, and Littman]{liabstractions}
L.~Li, T.~J. Walsh, and M.~L. Littman.
\newblock Towards a unified theory of state abstraction for mdps.
\newblock In \emph{In Proceedings of the Ninth International Symposium on
  Artificial Intelligence and Mathematics}, pages 531--539, 2006.

\bibitem[Konidaris and Barto(2009)]{konidarisabstractions}
G.~Konidaris and A.~Barto.
\newblock Efficient skill learning using abstraction selection.
\newblock In \emph{Twenty-First International Joint Conference on Artificial
  Intelligence}, 2009.

\bibitem[Browne et~al.(2012)Browne, Powley, Whitehouse, Lucas, Cowling,
  Rohlfshagen, Tavener, Perez, Samothrakis, and Colton]{mcts}
C.~B. Browne, E.~Powley, D.~Whitehouse, S.~M. Lucas, P.~I. Cowling,
  P.~Rohlfshagen, S.~Tavener, D.~Perez, S.~Samothrakis, and S.~Colton.
\newblock A survey of monte carlo tree search methods.
\newblock \emph{IEEE Transactions on Computational Intelligence and AI in
  games}, 4\penalty0 (1):\penalty0 1--43, 2012.

\bibitem[Helmert(2006)]{fd}
M.~Helmert.
\newblock The fast downward planning system.
\newblock \emph{Journal of Artificial Intelligence Research}, 26:\penalty0
  191--246, 2006.

\bibitem[Srivastava et~al.(2014)Srivastava, Fang, Riano, Chitnis, Russell, and
  Abbeel]{srivastava2014combined}
S.~Srivastava, E.~Fang, L.~Riano, R.~Chitnis, S.~Russell, and P.~Abbeel.
\newblock Combined task and motion planning through an extensible
  planner-independent interface layer.
\newblock In \emph{2014 IEEE international conference on robotics and
  automation (ICRA)}, pages 639--646. IEEE, 2014.

\bibitem[Moerland et~al.(2020)Moerland, Deichler, Baldi, Broekens, and
  Jonker]{moerl2020think}
T.~M. Moerland, A.~Deichler, S.~Baldi, J.~Broekens, and C.~M. Jonker.
\newblock Think too fast nor too slow: The computational trade-off between
  planning and reinforcement learning, 2020.

\bibitem[Stilman and Kuffner(2005)]{namo}
M.~Stilman and J.~J. Kuffner.
\newblock Navigation among movable obstacles: Real-time reasoning in complex
  environments.
\newblock \emph{International Journal of Humanoid Robotics}, 2\penalty0
  (04):\penalty0 479--503, 2005.

\bibitem[Guestrin et~al.(2003)Guestrin, Koller, Parr, and
  Venkataraman]{guestrin2003efficient}
C.~Guestrin, D.~Koller, R.~Parr, and S.~Venkataraman.
\newblock Efficient solution algorithms for factored mdps.
\newblock \emph{Journal of Artificial Intelligence Research}, 19:\penalty0
  399--468, 2003.

\bibitem[Baum et~al.(2012)Baum, Nicholson, and Dix]{baum2012proximity}
J.~Baum, A.~E. Nicholson, and T.~I. Dix.
\newblock Proximity-based non-uniform abstractions for approximate planning.
\newblock \emph{Journal of Artificial Intelligence Research}, 43:\penalty0
  477--522, 2012.

\bibitem[Hernandez{-}Gardiol(2008)]{gardiolthesis}
N.~Hernandez{-}Gardiol.
\newblock \emph{Relational envelope-based planning}.
\newblock PhD thesis, Massachusetts Institute of Technology, Cambridge, MA,
  {USA}, 2008.
\newblock URL \url{http://hdl.handle.net/1721.1/43028}.

\bibitem[Boutilier(1997)]{boutilier1997correlated}
C.~Boutilier.
\newblock Correlated action effects in decision theoretic regression.
\newblock In \emph{UAI}, pages 30--37, 1997.

\bibitem[Abel et~al.(2019)Abel, Umbanhowar, Khetarpal, Arumugam, Precup, and
  Littman]{abel2019value}
D.~Abel, N.~Umbanhowar, K.~Khetarpal, D.~Arumugam, D.~Precup, and M.~L.
  Littman.
\newblock Value preserving state-action abstractions, 2019.

\bibitem[Driess et~al.(2020)Driess, Oguz, Ha, and Toussaint]{driess2020deep}
D.~Driess, O.~Oguz, J.-S. Ha, and M.~Toussaint.
\newblock Deep visual heuristics: Learning feasibility of mixed-integer
  programs for manipulation planning.
\newblock In \emph{IEEE International Conference on Robotics and Automation
  (ICRA)}, 2020.

\bibitem[Carpentier et~al.(2017)Carpentier, Budhiraja, and
  Mansard]{carpentier2017learning}
J.~Carpentier, R.~Budhiraja, and N.~Mansard.
\newblock Learning feasibility constraints for multi-contact locomotion of
  legged robots.
\newblock In \emph{Robotics: Science and Systems}, page~9p, 2017.

\bibitem[McDermott et~al.(1998)McDermott, Ghallab, Howe, Knoblock, Ram, Veloso,
  Weld, and Wilkins]{pddl}
D.~McDermott, M.~Ghallab, A.~Howe, C.~Knoblock, A.~Ram, M.~Veloso, D.~Weld, and
  D.~Wilkins.
\newblock {PDDL}-the planning domain definition language, 1998.

\bibitem[Coumans and Bai(2016)]{pybullet}
E.~Coumans and Y.~Bai.
\newblock Pybullet, a python module for physics simulation for games, robotics
  and machine learning.
\newblock \emph{GitHub repository}, 2016.

\bibitem[Silver et~al.(2016)Silver, Huang, Maddison, Guez, Sifre, van~den
  Driessche, Schrittwieser, Antonoglou, Panneershelvam, Lanctot, Dieleman,
  Grewe, Nham, Kalchbrenner, Sutskever, Lillicrap, Leach, Kavukcuoglu, Graepel,
  and Hassabis]{alphago}
D.~Silver, A.~Huang, C.~Maddison, A.~Guez, L.~Sifre, G.~van~den Driessche,
  J.~Schrittwieser, I.~Antonoglou, V.~Panneershelvam, M.~Lanctot, S.~Dieleman,
  D.~Grewe, J.~Nham, N.~Kalchbrenner, I.~Sutskever, T.~Lillicrap, M.~Leach,
  K.~Kavukcuoglu, T.~Graepel, and D.~Hassabis.
\newblock Mastering the game of {G}o with deep neural networks and tree search.
\newblock \emph{Nature}, 2016.

\bibitem[Kim et~al.(2018)Kim, Kaelbling, and Lozano-P{\'e}rez]{ltampkim2}
B.~Kim, L.~P. Kaelbling, and T.~Lozano-P{\'e}rez.
\newblock Guiding search in continuous state-action spaces by learning an
  action sampler from off-target search experience.
\newblock In \emph{Thirty-Second AAAI Conference on Artificial Intelligence},
  2018.

\bibitem[Kim and Shimanuki(2019)]{kim19learning}
B.~Kim and L.~Shimanuki.
\newblock Learning value functions with relational state representations for
  guiding task-and-motion planning.
\newblock \emph{Conference on Robot Learning}, 2019.

\bibitem[Chitnis et~al.(2016)Chitnis, Hadfield-Menell, Gupta, Srivastava,
  Groshev, Lin, and Abbeel]{ltampchitnis}
R.~Chitnis, D.~Hadfield-Menell, A.~Gupta, S.~Srivastava, E.~Groshev, C.~Lin,
  and P.~Abbeel.
\newblock Guided search for task and motion plans using learned heuristics.
\newblock In \emph{Robotics and Automation (ICRA), 2016 IEEE International
  Conference on}, pages 447--454. IEEE, 2016.

\bibitem[Jong and Stone(2005)]{jong2005state}
N.~K. Jong and P.~Stone.
\newblock State abstraction discovery from irrelevant state variables.
\newblock In \emph{IJCAI}, volume~8, pages 752--757, 2005.

\bibitem[Steinkraus(2005)]{steinkraus2005solving}
K.~A. Steinkraus.
\newblock \emph{Solving large stochastic planning problems using multiple
  dynamic abstractions}.
\newblock PhD thesis, Massachusetts Institute of Technology, 2005.

\bibitem[Abel et~al.(2017)Abel, Hershkowitz, and Littman]{abel2017near}
D.~Abel, D.~E. Hershkowitz, and M.~L. Littman.
\newblock Near optimal behavior via approximate state abstraction.
\newblock \emph{arXiv preprint arXiv:1701.04113}, 2017.

\bibitem[Bean et~al.(1987)Bean, Birge, and Smith]{bean1987aggregation}
J.~C. Bean, J.~R. Birge, and R.~L. Smith.
\newblock Aggregation in dynamic programming.
\newblock \emph{Operations Research}, 35\penalty0 (2):\penalty0 215--220, 1987.

\bibitem[Bertsekas et~al.(1988)Bertsekas, Castanon,
  et~al.]{bertsekas1988adaptive}
D.~P. Bertsekas, D.~A. Castanon, et~al.
\newblock Adaptive aggregation methods for infinite horizon dynamic
  programming.
\newblock \emph{IEEE Transactions on Automatic Control}, 1988.

\bibitem[Jiang et~al.(2015)Jiang, Kulesza, and Singh]{jiang2015abstraction}
N.~Jiang, A.~Kulesza, and S.~Singh.
\newblock Abstraction selection in model-based reinforcement learning.
\newblock In \emph{International Conference on Machine Learning}, pages
  179--188, 2015.

\bibitem[Konidaris(2016)]{konidaris2016constructing}
G.~Konidaris.
\newblock Constructing abstraction hierarchies using a skill-symbol loop.
\newblock In \emph{IJCAI: proceedings of the conference}, volume 2016, page
  1648. NIH Public Access, 2016.

\bibitem[Zhang and Poole(1999)]{zhang1999role}
N.~L. Zhang and D.~Poole.
\newblock On the role of context-specific independence in probabilistic
  inference.
\newblock In \emph{IJCAI}, volume~1, page~9, 1999.

\bibitem[Poole(2013)]{poole2013context}
D.~L. Poole.
\newblock Context-specific approximation in probabilistic inference.
\newblock \emph{arXiv preprint arXiv:1301.7408}, 2013.

\bibitem[Domshlak and Shimony(2004)]{domshlak2004efficient}
C.~Domshlak and S.~E. Shimony.
\newblock Efficient probabilistic reasoning in bns with mutual exclusion and
  context-specific independence.
\newblock \emph{International journal of intelligent systems}, 19\penalty0
  (8):\penalty0 703--725, 2004.

\bibitem[Hoey et~al.(1999)Hoey, St-Aubin, Hu, and Boutilier]{spudd}
J.~Hoey, R.~St-Aubin, A.~J. Hu, and C.~Boutilier.
\newblock Spudd: Stochastic planning using decision diagrams.
\newblock In \emph{UAI}, 1999.

\bibitem[Boutilier et~al.(1999)Boutilier, Dean, and
  Hanks]{boutilier1999decision}
C.~Boutilier, T.~Dean, and S.~Hanks.
\newblock Decision-theoretic planning: Structural assumptions and computational
  leverage.
\newblock \emph{Journal of Artificial Intelligence Research}, 11:\penalty0
  1--94, 1999.

\bibitem[Kuffner and LaValle(2000)]{rrtconnect}
J.~J. Kuffner and S.~M. LaValle.
\newblock Rrt-connect: An efficient approach to single-query path planning.
\newblock In \emph{Proceedings 2000 ICRA. Millennium Conference. IEEE
  International Conference on Robotics and Automation. Symposia Proceedings
  (Cat. No. 00CH37065)}, volume~2, pages 995--1001. IEEE, 2000.

\bibitem[Hoffmann(2001)]{ff}
J.~Hoffmann.
\newblock {FF}: The fast-forward planning system.
\newblock \emph{AI magazine}, 22\penalty0 (3):\penalty0 57--57, 2001.

\bibitem[Kingma and Ba(2014)]{adam}
D.~P. Kingma and J.~Ba.
\newblock Adam: A method for stochastic optimization.
\newblock \emph{arXiv preprint arXiv:1412.6980}, 2014.

\end{thebibliography}

\appendix
\newpage

\section{Pseudocode: Approximate Context-Specific Independence Learning}
\label{sec:csipseudocode}
The following pseudocode describes our algorithm for learning approximate context-specific independences (CSIs). See Figure \ref{fig:campexample}B for an example output, and see \secref{subsec:csilearning} for discussion.

\begin{algorithm}[H]
  \SetAlgoNoEnd
  \DontPrintSemicolon
  \SetKwFunction{algo}{algo}\SetKwFunction{proc}{proc}
  \SetKwProg{myalg}{Algorithm}{}{}
  \SetKwProg{myproc}{Subroutine}{}{}
  \SetKw{Continue}{continue}
  \SetKw{Break}{break}
  \SetKw{Return}{return}
    \nonl \textbf{Input}: State and action variables $V =\{S^1, \ldots, S^n\} \cup \{A^1, \ldots, A^m\}$\;
    \nonl \textbf{Input}: Black-box transition model $T$\;
    \nonl \textbf{Input}: Context $(C, \C)$\;
    \nonl \textbf{Input}: Number of samples $k_1, k_2$ \quad\tcp{\footnotesize Hyperparameters}
    \nonl \textbf{Returns}: Approximate CSIs $\{(V^i, V^j) : V^j_{t+1} \indep  V^i_{t} \mid (C, \C) \}$, for arbitrary $t$\;
    \tcp{\footnotesize Initialize all pairs of variables to be independent}
    \nonl \textbf{Initialize}: \textsc{CSIs} $ \gets V \times V$\;
    \tcp{\footnotesize Sample $k_1$ state and action assignments in the context}
    \nonl $U \gets \textsc{SampleInContext}(V, C, \C, k_1)$\;
    \tcp{\footnotesize Test pairs of variables for dependence}
    \nonl \For{$V^i, V^j \in V \setminus C$}
    {
    \nonl \For{$u \in U$}
    {
    \nonl \For{up to $k_2$ samples $v^i$ of $V^i$}
    {
     \nonl \If{$T(V_{t+1}^j \mid V_t = u) \neq T(V_{t+1}^j \mid V_t\setminus \{V_t^i\} = u_{-i}, V_t^i = v^i) $}
     {
     \tcp{\footnotesize $V^j$ is dependent on $V^i$; remove this pair from CSIs}
     $\textsc{CSIs} \gets \textsc{CSIs} \setminus \{(V^i, V^j)\}$
     }
    }
    }
    }
     \Return{\textsc{CSIs}}
\end{algorithm}

\section{Pseudocode: Context Selector Learning}
\label{sec:selectorpseudocode}
The following pseudocode describes our algorithm for learning a context selector model, given training tasks and their context-specific independences (CSIs). See \secref{subsec:selectorlearning} for discussion.

\begin{algorithm}[H]
  \SetAlgoNoEnd
  \DontPrintSemicolon
  \SetKwFunction{algo}{algo}\SetKwFunction{proc}{proc}
  \SetKwProg{myalg}{Algorithm}{}{}
  \SetKwProg{myproc}{Subroutine}{}{}
  \SetKw{Continue}{continue}
  \SetKw{Break}{break}
  \SetKw{Return}{return}
    \nonl \textbf{Input}: Training tasks $W_{\text{train}} = \{\omega^{(i)}\}_{i=1}^{N}$ with features $\{\theta_{\omega^{(i)}} \}_{i=1}^{N}$\;
    \nonl \textbf{Input}: Black-box transition model $T$\;
    \nonl \textbf{Input}: Set of contexts $\{(C, \C) \}$\;
    \nonl \textbf{Input}: All learned \textsc{CSIs} $(C, \C) \to \{(V^i, V^j) : V^j_{t+1} \indep  V^i_{t} \mid (C, \C) \}$\;
    \nonl \textbf{Returns}: Context selector $f_\alpha(\theta_\omega) = (C, \tilde{\mathcal{C}})^*$ \quad\tcp{\footnotesize Neural network with parameters $\alpha$}
    \nonl \textbf{Initialize}: Inputs for supervised learning $X \gets [\theta_{\omega^{(1)}}, \ldots, \theta_{\omega^{(N)}}]$\;
    \nonl \textbf{Initialize}: Targets for supervised learning $Y \gets []$\;
    \nonl \For{$\omega^{(i)} \in W_{\text{train}}$}
    {
     \tcp{\footnotesize See Subroutine below}
    $Y[i] \gets \argmax_{(C, \C)} \textsc{ScoreContext}(\omega^{(i)}, (C, \C), T, \textsc{CSIs} \text{ for } (C, \C))$
    }
    \tcp{\footnotesize Perform supervised learning (multiclass classification)}
    \nonl $\alpha^* \gets \argmin_{\alpha} \textsc{CrossEntropyLoss}(X, Y; \alpha)$\;
    \nonl \Return{$f_{\alpha^*}$}\;
    \;
    \myproc{\textsc{ScoreContext}}{
    \nonl \textbf{Input}: Training task $\omega = (s_0, R)$\;
    \nonl \textbf{Input}: Black-box transition model $T$\; 
    \nonl \textbf{Input}: Context $(C, \C)$\;
    \nonl \textbf{Input}: Learned \textsc{CSIs} $\{(V^i, V^j) : V^j_{t+1} \indep  V^i_{t} \mid (C, \C) \}$\;
    \nonl \textbf{Returns}: A score\;
    \nonl $\M' \gets \textsc{CreateCAMP}(T, R, (C, \C), \textsc{CSIs})$ \quad\tcp{\footnotesize See \secref{sec:representation}}
    \nonl $\pi(s) \triangleq \plan{}(\M', s)$ \quad\tcp{Plan in the \camp{}}
    \nonl \Return{$J(\pi, \omega)$} \quad\tcp{\footnotesize See Equation \ref{eq:objective}}
    }
\end{algorithm}

\section{\camp{} Graphical Example}
\label{sec:campgraphicalexample}
\begin{figure}[t]
\includegraphics[width=\textwidth]{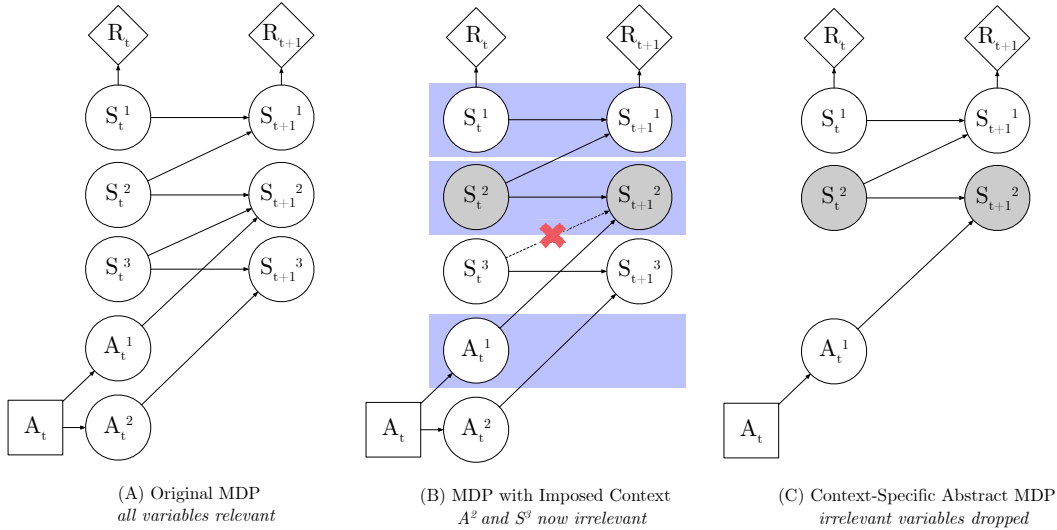}
\centering
\caption{
(A) Example of a factored \mdp{} represented as an influence diagram \cite{boutilier1999decision}.
As seen in the diagram, $S_{\text{rew}} = \{ S^1 \}$.
With no contexts imposed, all variables are relevant.
(B) Imposing contexts can induce new independences. In this example, a context involving $S^2$ is imposed, inducing an independence between $S^2$ and $S^3$ (red $\times$). Variables $A^2$ and $S^3$ are irrelevant under the imposed context; relevant variables are highlighted in blue. Note that relevance is a time-independent property.
(C) Dropping the irrelevant variables leads to a \camp{}, an abstraction of the original \mdp{}.
}
\label{fig:campexample}
\end{figure}

\figref{fig:campexample} provides an example of a \camp{}. Note that standard influence diagrams~\cite{boutilier1999decision} cannot capture context-specific independence, so we use a dotted line in the second panel to denote this concept.

\section{Domain and Planner Details}
\label{sec:domplandetails}

\textbf{Domain D1: Gridworld.} The first domain we consider is a simple maze-style gridworld in which the agent must navigate across rooms to reach a goal location, while avoiding obstacles that stochastically move around at each timestep. The agent has available to it \textsc{remove(obj)} actions, which remove the given obstacle from the world so that the agent can no longer collide with it, but these actions can only be used when the agent is adjacent to the obstacle. Whenever the agent collides with an obstacle, it is placed back in its initial location. Each obstacle remains within a particular room, and so the agent can impose a context of not entering particular rooms, allowing it to ignore the obstacles that are in those rooms, and also not have to consider the action of removing those obstacles. Across tasks, we vary the maze layout.
We train on 50 task instances and test on 10 held-out instances.

\emph{Planners.} We consider the following planners for this domain: Monte Carlo tree search (MCTS), breadth-first graph search with replanning (BFSReplan), and asynchronous value iteration (VI). Both MCTS and BFSReplan are online planners, while VI is offline. As such, VI computes a policy over the full state space, and thus is only tractable in this relatively small (about 100,000 states) domain.

\emph{Representations.} The features of each task are a top-down image of the maze layout. The state is a vector of the current position and room of each obstacle, the agent, and the goal. The actions are moving up, down, left, right; and removing each obstacle in the environment.

\textbf{Domain D2: Classical planning.}
We next consider a deterministic classical planning domain in which an agent must make a meal for dinner, and has three options: to stay within the living room to make ramen, to go to the kitchen to make a sandwich, or to go to the store to buy and prepare a steak. Making any of these terminates the task. The steak gives higher terminal reward than the sandwich, which in turn gives higher terminal reward than the ramen. However, planning to go to the store for steak requires reasoning about many objects that would be irrelevant under the context of staying within the home (for a sandwich or ramen), and planning to go to the kitchen for a sandwich requires reasoning about many objects that would be irrelevant under the context of staying within the living room (for ramen).
There is also a timestep penalty, incentivizing the agent to finish quickly.
Optimal plans may involve 2, 16, or 22 actions depending on the relative rewards for obtaining the ramen, sandwich, and steak.
These rewards are the only thing that varies between task instances; there is thus small variation between task instances relative to the other domains.
We train on 20 task instances and test on 25 held-out instances.

\emph{Planner.} We use an off-the-shelf classical planner (Fast-Downward~\cite{fd} in $\text{A}^*$ mode with the \emph{lmcut} heuristic). The various rewards are implemented as action costs. As this domain is deterministic, we only run the planner once per task; it is guaranteed to find a reward-maximizing trajectory.

\emph{Representations.} The features of each task are a vector of the terminal rewards for each meal. The state is a binary vector describing which logical fluents hold true (1) versus false (0). The actions are logical operators described in PDDL, each containing parameters, preconditions, and effects.

We also use this domain as a testbed for additional experiments into the impact of $\lambda$ (the trade-off parameter in \eqnref{eq:objective}), and the number of training tasks on our method. See \secref{subsec:lambda} and \secref{subsec:numenvs}.

\textbf{Domain D3: Robotic navigation among movable obstacles (\namo{}).} Illustrated in \figref{fig:teaser}A, this domain has a robot navigating through rooms with the goal of reaching the red object in the upper-right room. Roughly 20 blue obstacles are scattered throughout the rooms, and like in the gridworld, the robot may impose the context of not entering particular rooms; it may also pick up the obstacles and move them out of its way. Across tasks, we vary the positions of all objects.
We train on 50 task instances and test on 10 held-out instances.
This domain has continuous states and actions, and as such is extremely challenging for planning. Though the obstacles do not move on their own (like they do in the gridworld), the difficulty of this domain stems from the added complexity of needing to reason about geometry and continuous trajectories. We simulate this domain using PyBullet~\cite{pybullet}. The reward function is sparse: 1000 if the goal location is reached and 0 otherwise.

\emph{Planner.} Developing planners for robotic domains with continuous states and actions is an active area of research. For this domain, we use a state-of-the-art task and motion planner~\cite{srivastava2014combined}, which is \emph{not} specific to \namo{} problems. We use the RRT-Connect algorithm~\cite{rrtconnect} for motion planning and the Fast-Forward PDDL planner~\cite{ff} for task planning.

\emph{Representations.} The features of each task are a top-down image of the scene. The state is a vector of the current pose of each object and the robot, and the robot's current room. The actions are moving the robot base to a target pose, and clearing an object in front of the robot.

\textbf{Domain D4: Robotic sequential manipulation.} Illustrated in \figref{fig:teaser}C, this domain has a robot manipulating the two red objects that start off on the left table to be placed into the bins on the right table. The fifteen blue objects on the left table serve as distractors, with which the robot must be careful not to collide when grasping the red objects; the green objects in the bins indicate that certain bins are already occupied. Across tasks, we vary the positions of all objects, and which bins are occupied by green objects. We also vary the radii of the red objects.
We train on 50 task instances and test on 10 held-out instances. We again simulate this domain using PyBullet~\cite{pybullet}. As in Domain D3, the reward function is sparse: 1000 if the goal location is reached and 0 otherwise.

Broadly, there are two types of contexts that are useful to impose in this domain. (1) If the robot chooses to constrain its \emph{grasp style} to only allow top-grasping the red objects, then it need not worry about colliding with the blue objects, and can thus ignore them. However, this does not always work, since not all geometries are amenable to being top-grasped; for instance, sometimes an object's radius may be too large. Note, however, that to place the red objects into the bins upright, a side-grasp is necessary, and so we provide the robot a \emph{regrasp} operator in addition to the standard move, pick, and place. Importantly, this regrasp operator is never \emph{necessary}, but including it can allow the robot to simplify its planning problem by ignoring the blue objects (see Equation \ref{eq:objective}). (2) If the robot chooses to constrain which bins it will place the red objects into, then it need not worry about the green objects in the other bins, simplifying the planning problem.

\emph{Planner.} Same as in Domain D3 (\namo{}).

\emph{Representations.} The features of each task are a vector of the object radii and occupied bins. The state is a vector of the current pose of each object, the grasp style used by the robot, and the current held object (if any). The actions are moving the robot to a target base pose and grasping at a target gripper pose (which requires an empty gripper), and moving the robot to a target base pose and placing at a target placement pose (which requires an object to be currently held).

\section{Experimental Details}
\label{subsec:details}

In all experiments, computational cost is measured in wall-clock time (seconds).
We use the following values of $\lambda$: 0 for MCTS\footnote{Since MCTS is an anytime algorithm, we give it a timeout of 0.25 seconds. With $\lambda=0$, the objective then reflects the best returns found within this timeout.}, 100 for BFSReplan, 250 for FastDownward, and 100 for \tamp{}.
Every domain uses horizon $H=25$. Additionally, to ensure that shorter plans are preferred in general, all domains use a discount factor $\gamma=0.99$, except for Domain D2 which uses a timestep penalty as previously discussed.

To properly evaluate our objective (\eqnref{eq:objective}), we would need to run every method until it completes, which can be extremely slow, e.g. for the pure planning baseline, or when our context selector picks a bad context. To safeguard against this, we impose a timeout of 60 seconds on all planning calls.

All neural networks are either fully connected for vector inputs or convolutional for image inputs. Fully connected networks have hidden layer sizes [50, 32, 10]. Convolutional networks use a convolutional layer with 10 output channels and kernel size 2, followed by a max-pooling layer with stride 2, and then fully connected layers of [32, 10]. Neural networks are trained using the Adam optimizer~\cite{adam} with learning rate $10^{-4}$, until the loss on the training dataset reaches $10^{-3}$.

To generate the spaces of contexts, we use the method described in \secref{subsec:csilearning}. In Domains D1, D2, and D3, we consider disjunctive and single-term (a single variable and a single value in its domain) constraints only, while in Domain D4 we also consider conjunctive constraints. All contexts only consider the discrete variables in the domain. Our parameters $k_1$ and $k_2$ (Appendix \ref{sec:csipseudocode}) are: $k_1=k_2=50$ for Domain D1, $k_1=k_2=40$ for Domain D2, and $k_1=k_2=25$ for Domains D3 and D4.

\section{Performance as a Function of Number of Training Tasks}
\label{subsec:numenvs}

The following plots illustrate how the objective value (left) and returns (right) accrued by the \camp{} policy vary as a function of the number of training tasks, in Domain D2 (classical planning):

\begin{figure}[h]
\includegraphics[width=\textwidth]{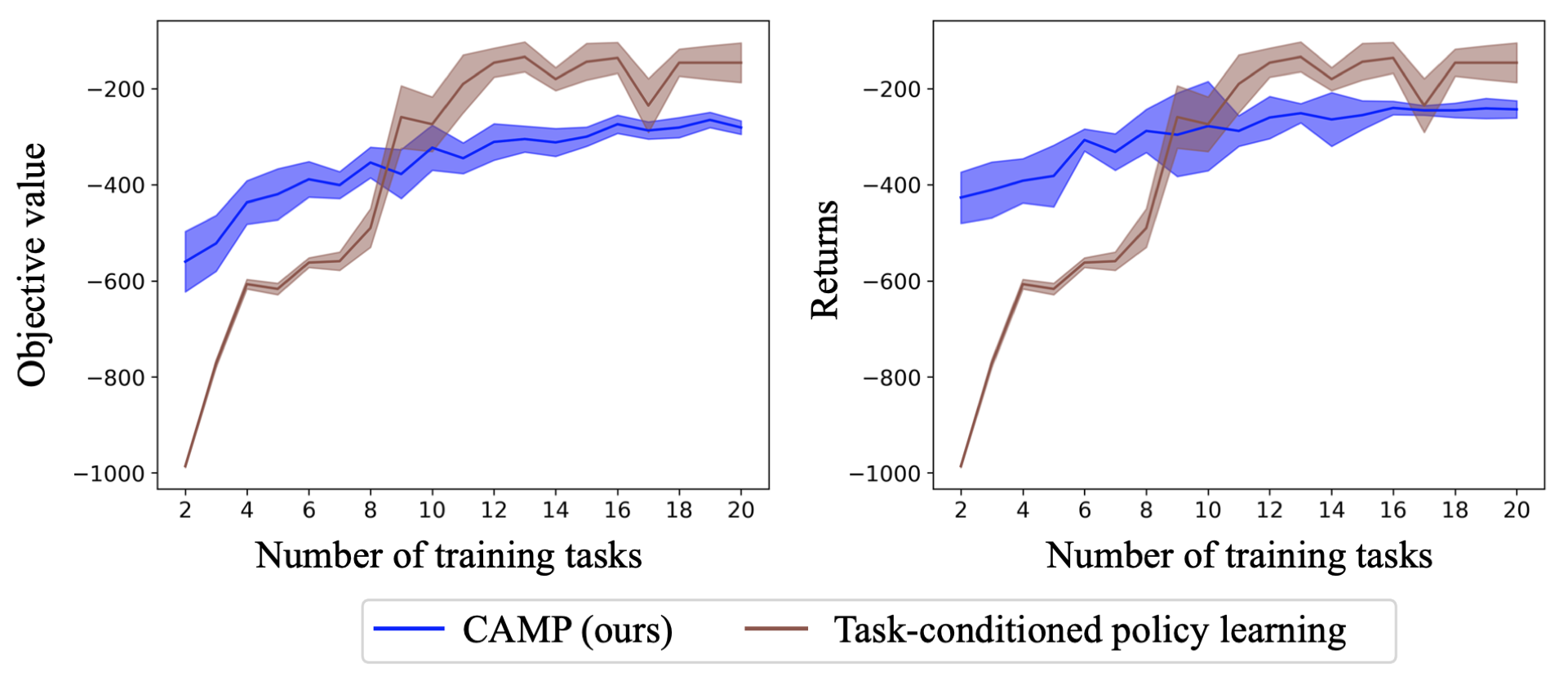}
\centering
\end{figure}

\emph{Discussion.} As following a policy in the test task requires near-zero computational effort (our neural networks are small enough that inference is very fast), the red curves in both plots are nearly identical. Interestingly, in the regime of fewer training tasks ($\leq 8$), \camp{} outperforms policy learning, despite policy learning performing better with the full set of 20 tasks. This leads us to believe that in domains where policy learning would perform well when given a lot of data, generating and planning in a \camp{} may be a more viable strategy when data is limited.
As shown in the main results, this disparity between \camp{}s and policy learning is more dramatic for the other three domains, where task instances are far more varied, so much so that \camp{}s sharply outperform policy learning for any reasonable number of training environments that we were able to test.

\section{Objective Values for Main Experiments}
\label{sec:objectivetable}

\begin{table}
\small
\centering
\begin{tabular}{l@{\hspace{0.2cm}} ccccc}
\toprule
\textbf{Method}  &
\multicolumn{5}{c}{\textbf{Test Task Objective Value (St. Dev.)}} \\
\cmidrule{2-6}
 & D1 (Grid), MCTS & D1 (Grid), BFSReplan & D2 (Classical) & D3 (\namo{}) & D4 (Manip) \\
\midrule
\camp{} (ours) & \textbf{70} (16) & \textbf{21} (10) & -286 (9.6) & \textbf{896} (63) & \textbf{744} (94) \\
\camp{} ablation & 25 (11) & 0.9 (24) & -308 (52) & 707 (154) & 453 (237) \\ 
Pure planning & 6 (5) & -17 (11) & -414 (20) & 242 (385) & 335 (86) \\ 
Plan transfer & -7 (0.4) & -6 (15) & -467 (0.02) & 141 (227) & 21 (34) \\ 
Policy learning & -3 (4) & -11 (13) & -469 (0.2) & -0.2 (0.01) & -0.2 (0.01) \\ 
Task-conditioned & 5 (5) & -2 (11) & \textbf{-145} (0.4) & -0.3 (0.01) & -0.2 (0.02) \\ 
Stilman's \cite{namo} & - & - & - & 826 (36) & - \\

\bottomrule
\end{tabular}
\vspace{0.2cm}
\caption{Compilation of test task objective values on all our domains and methods. Objective values are computed using the same values of $\lambda$ that were used during training. All table entries report an average over 10 independent runs of both context selector training and test task evaluation. Stilman's algorithm~\cite{namo} is \namo{}-specific and so is only run on the \namo{} domain.}
\label{tab:mainresults}

\end{table}

Table \ref{tab:mainresults} complements \figref{fig:plots2d} in the main text, showing the objective values obtained for all domains, planners, and methods with the values of $\lambda$ that were used during training.
See \secref{subsec:results} for further analysis and discussion.

\section{Performance with an Offline Planner}
\label{subsec:offline}

The table on the right shows test task objective values with an offline planner (asynchronous value iteration), in Domain D1 (gridworld).

\begin{wraptable}{r}{-40mm}
\small
\centering
\begin{tabular}{l@{\hspace{0.2cm}} c}
\toprule
\textbf{Method}  & \textbf{Objective (SD)} \\
\midrule


\camp{} (ours) & \textbf{25} (3) \\ 
\camp{} ablation & 4 (11) \\ 
Pure planning & -1 (2) \\ 
Plan transfer & - \\ 
Policy learning & 7 (0.01) \\ 
Task-conditioned & 7 (0.01) \\ 





\bottomrule
\end{tabular}

\end{wraptable}

\emph{Discussion.}
This result mirrors the trend found in the main results: \camp{}s strongly outperform both pure policy learning and pure planning, for reasons of generalization error and high computational cost, respectively.
\camp{}'s reduction of the state space leads to substantial benefits for offline planning, because offline planners find a policy over the entire state space.
Nonetheless, \camp{} remains primarily motivated by online planning for robotics, where the continuous states and actions make offline planning completely infeasible in practice. 

\end{document}